\documentclass[letterpaper]{article}
\usepackage{proceed2e}
\usepackage[margin=1in]{geometry}

\usepackage{bbm}
\usepackage{times}
\usepackage{amsmath}
\usepackage{amsthm}
\usepackage{amssymb}
\usepackage{graphicx}
\usepackage{algorithm}
\usepackage{algorithmic}
\usepackage{caption}
\usepackage{subcaption}
\newtheorem{theorem}{Theorem}
\newtheorem{corollary}{Corollary}
\newtheorem{remark}{Remark}
\newtheorem{definition}{Definition}
\newtheorem{proposition}{Proposition}
\newtheorem{lemma}{Lemma}

\newcommand{\balpha}{\boldsymbol{\alpha}}
\newcommand{\bdelta}{\boldsymbol{\Delta}}
\newcommand{\bg}{\boldsymbol{g}}
\newcommand{\bh}{\boldsymbol{h}}
\newcommand{\bpi}{\boldsymbol{\pi}}
\newcommand{\bG}{\boldsymbol{G}}
\newcommand{\bY}{\boldsymbol{Y}}
\newcommand{\cK}{\mathcal{K}}
\newcommand{\cY}{\mathcal{Y}}
\newcommand{\cP}{\mathcal{P}}
\newcommand{\cE}{\mathcal{E}}
\newcommand{\cF}{\mathcal{F}}

\usepackage{natbib}

\usepackage{color}

\title{Analysis of Thompson Sampling for Graphical Bandits Without the Graphs}

\author{} 

\author{Fang Liu \\
The Ohio State University\\
Columbus, Ohio 43210\\
liu.3977@osu.edu\\
\And Zizhan Zheng\\
Tulane University\\
New Orleans, LA 70118\\
zzheng3@tulane.edu\\
\And Ness Shroff\\
The Ohio State University\\
Columbus, Ohio 43210\\
shroff.11@osu.edu\\
}

\begin{document}

\maketitle

\begin{abstract}
We study multi-armed bandit problems with graph feedback, in which the decision maker is allowed to observe the neighboring actions of the chosen action, in a setting where the graph may vary over time and is never fully revealed to the decision maker. We show that when the feedback graphs are undirected, the original Thompson Sampling achieves the optimal (within logarithmic factors) regret $\tilde{O}\left(\sqrt{\beta_0(G)T}\right)$ over time horizon $T$, where $\beta_0(G)$ is the average independence number of the latent graphs. To the best of our knowledge, this is the first result showing that the original Thompson Sampling is optimal for graphical bandits in the undirected setting. A slightly weaker regret bound of Thompson Sampling in the directed setting is also presented. To fill this gap, we propose a variant of Thompson Sampling, that attains the optimal regret in the directed setting within a logarithmic factor. Both algorithms can be implemented efficiently and do not require the knowledge of the feedback graphs at any time.
\end{abstract}

\section{Introduction}
\begin{table*}[h]
\caption{Comparison of the existing algorithms}
\label{table:comparison}
\begin{center}
\begin{tabular}{|c|c|c|c|c|}
\hline
Algorithm & Reference & Graph & Undirected & Directed\\\hline
\multicolumn{5}{|c|}{Non-stochastic graphical bandits}\\\hline
ExpBan & \cite{mannor} & Informed & \multicolumn{2}{c|}{$O\left(\sqrt{\chi(G)T\log K}\right)$}\\\hline
ELP & \cite{mannor} & Informed & $O\left(\sqrt{\beta_0(G)T\log K}\right)$ & $O\left(\sqrt{\chi(G)T\log K}\right)$ \\\hline
Exp3-SET & \cite{alon2013bandits} & Uninformed & $O\left(\sqrt{\beta_0(G)T\log K}\right)$ & $O\left(\sqrt{mas(G)T\log K}\right)$ \\\hline
Exp3-DOM & \cite{alon2013bandits} &Informed & \multicolumn{2}{c|}{$O\left(\sqrt{\beta_0(G)T\log (KT)}\log K\right)$}\\\hline
Exp3.G & \cite{alon2015online} & Uninformed & \multicolumn{2}{c|}{$O\left(\sqrt{\beta_0(G)T}\log (KT)\right)$}\\\hline
Exp3-IX & \cite{kocak2014efficient} & Uninformed & \multicolumn{2}{c|}{$O\left(\sqrt{\beta_0(G)T\log K\log (KT)}\right)$}\\\hline
\multicolumn{5}{|c|}{Stochastic graphical bandits}\\\hline
Cohen's Algo. 1 & \cite{Cohen2016} & Without & \multicolumn{2}{c|}{$O\left(\sqrt{\beta_0(G)T\log K}\log (KT)\right)$}\\\hline
IDS-N/IDSN-LP & \cite{liu2017information} & Informed & \multicolumn{2}{c|}{$O\left(\sqrt{\chi(G)T\log K}\right)$}\\\hline
TS-N & \cite{liu2017information} & Without & \multicolumn{2}{c|}{$O\left(\sqrt{\chi(G)T\log K}\right)$}\\\hline
TS-N & this paper & Without & $O\left(\sqrt{\beta_0(G)T\log K}\right)$ & $O\left(\sqrt{mas(G)T\log K}\right)$ \\\hline
TS-U & this paper & Without & \multicolumn{2}{c|}{$O\left(\sqrt{\beta_0(G)T\log K\log (KT)}\right)$}\\\hline
\end{tabular}
\end{center}
\end{table*}

Multi-Armed Bandits (MAB) models are quintessential models for sequential decision making. In the classical MAB setting, at each time, a policy must choose an action from a set of $K$ actions with unknown probability distributions. Choosing an action $i$ at time $t$ reveals a random reward $X_{i}(t)$ drawn from the probability distribution of action $i.$  The goal is to find policies that minimize the expected loss due to uncertainty about actions' distributions over a given time horizon $T$.      

In this work, we consider an important variant of bandit problems, called graphical bandits, where choosing an action $i$ not only generates a reward from action $i$, but also reveals observations for a subset of the remaining actions. Graphical bandits are also known as bandits with graph-structured feedback or bandits with side-observations, in which the feedback model is specified by a sequence $\{G_t\}_{t\geq1}$ of feedback graphs. Each feedback graph $G_t$ is a directed graph whose nodes correspond to the actions. An arc\footnote{We also use the notation $i\to j$ to represent an arc from node $i$ to node $j$ for simplicity.} $(i,j)$ in the graph indicates that the agent observes the reward of action $j$ if action $i$ is chosen in that round.

Motivating examples for situations where side observations are available include viral marketing and online pricing. Consider the viral marketing problem, where a decision maker wants to find the user with the maximum influence in an online social network (e.g., Facebook) to offer a promotion~(\cite{carpentier2016revealing}). Each time the decision maker offers a promotion to a user, it also has an opportunity to survey the user's neighbors in the network regarding their potential interest in the same offer. This is possible when the online network has an additional survey feature that generates ``side observations''. For example, when user $i$ is offered a promotion, her neighbors may be queried as follows: ``User $i$ was recently offered a promotion. Would you also be interested in the offer?". Here, choosing an action in the graphical bandit problem corresponds to choosing a user in the network and side-observations across actions are captured by the links in the social network. 

Consider another example in the online pricing problem, where a seller is selling goods on the Internet. In each round, the seller announces a price for the product. Then, a buyer arrives and decides whether or not to purchase the product based on its private value. A purchase takes place if and only if the announced price is no more than its private value. At the end of the round, the seller observes whether or not the buyer purchased the product at the announced price. If the buyer purchases the product, then the seller knows that the buyer would have bought the product at any lower price. Otherwise, the seller knows that the buyer would not have bought the product at any higher price. Here, actions in the graphical bandit problem corresponds to the prices that the seller can choose. The feedback graph is a directed graph over the prices that a price is connected to a lower (higher) price if and only if they are both below (above) the private value of the buyer.

Graphical bandits have been studied in both non-stochastic (adversarial) domain by~\cite{mannor,alon2013bandits,alon2015online,kocak2014efficient}, and stochastic domain by~\cite{bhagat,sigmetrics2014,buccapatnam2017reward,tossou2017thompson,liu2017information}. Regret bounds as a function of combinatorial properties of the feedback graphs are characterized in different settings: undirected graphs vs directed graphs, time-invariant graphs vs time-variant graphs.

However, most of the existing works mentioned above require prior knowledge of the feedback graphs for their algorithms to run. These algorithms fall into either the informed setting (where the algorithms have access to the graph structure before making decisions) or the uninformed setting (where the algorithms have access to the graph for performing their updates after decisions). 

The assumption that the feedback graph is disclosed to the decision maker does not hold in many real-world applications. For example, in the viral marketing problem, the third-party decision maker is not allowed to have the knowledge of the social network in order to protect the privacy of the users. In the online pricing problem, the private value of the buyer is never revealed to the seller. Thus the feedback graph is never disclosed to the seller. This motivates us to study the graphical bandits in a setting with limited information, where the feedback graphs are \emph{never fully revealed} to the decision maker.

In this work, we study the graphical bandits without the graphs in a general setting, where the graphs are allowed to be time-variant and directed. Moreover, the only feedback available to the decision maker at the end of each round is the out-neighborhood of the chosen action in the latent graph, along with the rewards associated with the observed actions. Generally speaking, our results show that Thompson Sampling algorithms (introduced by~\cite{thompson}) can achieve a regret bound of the form $\tilde{O}\left(\sqrt{\beta_0(G)T}\right)$, where $\beta_0(G)$ is the average independence number\footnote{See Section~\ref{subsec:graphmodel} for a brief review of the combinatorial properties of graphs.} of the latent graphs, that is optimal within logarithmic factors. More specifically, we make the following contributions to graphical bandits without knowledge of the feedback graphs. (Table~\ref{table:comparison} summarizes the main results.)
\begin{itemize}
\item We develop a problem-independent Bayesian regret  bound for the vanilla Thompson Sampling algorithm (TS-N) for graphical bandits without the graphs. In the undirected setting, where the latent graphs are undirected, we show that TS-N obtains the optimal (within logarithmic factors) regret bound of $O\left(\sqrt{\beta_0(G)T\log K}\right)$, where $\beta_0(G)$ is the average independence number of the latent graphs (Corollary~\ref{cor:TSNundirected}). Our regret bound is much sharper than the form of $O\left(\sqrt{\chi(G)T\log K}\right)$ that was shown by~\cite{liu2017information}, where $\chi(G)$ is the average clique cover number of the latent graphs, as $\beta_0(G)\leq \chi(G)$ in general. As far as we know, this is the first result showing that Thompson Sampling, without knowledge of the graph, can attain the optimal regret within logarithmic factors.
\item In the directed setting, where the graphs are allowed to be directed, we show that TS-N achieves $O\left(\sqrt{mas(G)T\log K}\right)$ regret in expectation, where $mas(G)$ is the average maximal acyclic subgraph number of the latent graphs (Corollary~\ref{cor:TSNdirected}). As a byproduct, our regret bounds for TS-N provide improved regret bounds for information directed sampling algorithms (IDS-N and IDSN-LP algorithms) proposed by~\cite{liu2017information}.
\item 
We propose a variant of the Thompson Sampling algorithm, TS-U, that achieves a regret bound of $O\left(\sqrt{\beta_0(G)T\log K\log (KT)}\right)$ for both the undirected and directed setting (Corollary~\ref{cor:TSU}). The regret bound of TS-U is optimal within logarithmic factors, and sharper than the state-of-the-art algorithm proposed by~\cite{Cohen2016}. Our results offer a recipe for practitioners to choose algorithms for graphical bandits without the graphs. If the latent graphs are known to be undirected, one can choose TS-N for the best regret guarantee. Otherwise, TS-U is the choice with the best guarantee.
\end{itemize}

\subsection{Related Work}
Graphical bandits were introduced in the non-stochastic domain by~\cite{mannor}. They propose the ExpBan algorithm that works in the time-invariant and informed setting, with the regret bound depending on the clique cover number. They also propose the ELP algorithm, that replaces the uniform distribution of Exp3 algorithm (proposed by~\cite{auer2002nonstochastic}) with a distribution that maximizes the minimum probability to observe an action. An optimal (within logarithmic factors) regret bound of ELP is shown in the undirected setting. 

However, the regret bound of the ELP depends on the clique cover number in the directed setting. These results are improved by~\cite{alon2013bandits}. They show that the vanilla Exp3 algorithm without mixing uniform distribution (Exp3-SET) achieves the same (but improved in the directed setting) regret bound as ELP, even in the uninformed setting. In the informed setting, they propose the Exp3-DOM algorithm, which is a variant of Exp3 algorithm with mixing uniform distribution over the dominating set of the feedback graph, achieves $\tilde{O}\left(\sqrt{\beta_0(G)T}\right)$ regret. This regret bound is further attained by Exp3.G~(\cite{alon2015online}) and Exp3-IX~(\cite{kocak2014efficient}) in the uninformed setting. The Exp3.G algorithm is a variant of Exp3-DOM where it replaces the dominating set with the universal set. The Exp3-IX algorithm uses a novel implicit exploration idea. However, these algorithms still require the knowledge of the feedback graphs for performing updates after the decisions in the uninformed setting.

Graphical bandits have also been considered in the stochastic domain by~\cite{bhagat}, who propose a natural variant of upper confidence bounds (introduced by~\cite{auer2002finite}) algorithm (UCB-N) and provide a problem-dependent regret guarantee depending on the clique cover number. This result is improved by \cite{sigmetrics2014} in the informed and time-invariant setting. Policies proposed by~\cite{sigmetrics2014}, namely $\epsilon_t$-greedy-LP and UCB-LP, are shown to be asymptotically optimal, both in terms of the graph structure and time. 

However, all of the afore-mentioned algorithms do not apply when the feedback graphs vary over the time and are never fully disclosed. Recently, researchers have developed new algorithms for graphical bandits in the setting with limited information, where the feedback graphs are time-variant, directed and never revealed to the decision maker. \cite{Cohen2016} propose an elimination-based algorithm that achieves the $\tilde{O}\left(\sqrt{\beta_0(G)T}\right)$ regret bound. \cite{tossou2017thompson} analyze the Bayesian regret performance of Thompson Sampling for the graphical bandits and provide a regret bound depending on the maximal clique cover number of the latent graphs. This result is improved to a regret bound depending on the average clique cover number by~\cite{liu2017information}. In this work, we provide sharper regret bounds for the vanilla Thompson Sampling (TS-N) and propose a variant of Thompson Sampling (TS-U) that obtains a better (within logarithmic factor) regret bound than the algorithm developed by~\cite{Cohen2016}.

Other related partial feedback models include label efficient bandit in~\cite{audibert2010regret} and prediction with limited advice in~\cite{seldin2014prediction}, where side observations are limited by a budget. Graphical bandits with Erd{\H{o}}s-R{\'e}nyi random graphs are studied by~\cite{kocak2016online,chen2016combinatorial,liu2017information}. Graphical bandits with noisy observations are studied by~\cite{kocak2016onlineb,wu2015online}.  A survey of the graphical bandits refers to~\cite{valko2016bandits}.

\section{Problem Formulation}
\subsection{Stochastic Bandit Model}
We consider a Bayesian formulation of the stochastic $K$-armed bandit problem in which uncertainties are modeled as random variables. At each time $t\in\mathbb{N}$, a decision maker chooses an action $A_t$ from a finite action set $\cK=\{1,\ldots,K\}$ and receives the corresponding random reward $Y_{t,A_t}$. Without loss of generality, we assume the space of possible rewards $\cY=[0,1]$. Note that the results in this work can be extended to the case where reward distributions are sub-Gaussian. There is a random variable $Y_{t,a}\in\cY$ associated with each action $a\in\cK$ and $t\in\mathbb{N}$. We assume that $\{Y_{t,a},\forall a\in\cK\}$ are independent for each time $t$. Let $\bY_t\triangleq(Y_{t,a})_{a\in\cK}$ be the vector of random variables at time $t\in\mathbb{N}$. The true reward distribution $p^*$ is a distribution over $\cY^K$, which is randomly drawn from the family of distributions $\cP$ and unknown to the decision maker. Conditioned on $p^*$, $(\bY_t)_{t\in\mathbb{N}}$ is an independent and identically distributed sequence with each element $\bY_t$ sampled from the distribution $p^*$. 

Let $A^*\in \arg\max_{a\in\cK}\mathbb{E}[Y_{t,a}|p^*]$ be the true optimal action conditioned on $p^*$. Then the $T$ period regret of the decision maker is the expected difference between the total rewards obtained by an oracle that always chooses the optimal action and the accumulated rewards up to time horizon $T$. Formally, we study the expected regret
\begin{equation}\label{eqn:regret}
\mathbb{E}[R(T)]=\mathbb{E}\left[\sum_{t=1}^TY_{t,A^*}-Y_{t,A_t}\right],
\end{equation}
where the expectation is taken over the randomness in the action sequence $(A_1,\ldots,A_T)$ and the outcomes $(\bY_t)_{t\in\mathbb{N}}$ and over the prior distribution over $p^*$. This notion of regret is also known as \emph{Bayesian regret}.

\subsection{Graph Feedback Model}\label{subsec:graphmodel}
In this problem, we assume the existence of side observations, which are described by a graph $G_t=(\cK,\cE_t)$ over the action set for each time $t$. The graph $G_t$ may be directed or undirected and can be dependent on time $t$. At each time $t$, the decision maker observes the reward $Y_{t,A_t}$ for playing action $A_t$ as well as the outcome $Y_{t,a}$ for each action $a\in\{a\in\cK|(A_t,a)\in\cE_t\}$. Note that it becomes the classical bandit feedback setting when the graph is empty (i.e., no edge exists) and it becomes the full-information (expert) setting when the graph is complete for all time $t$. Note that the graph $G_t$ is {\it never fully revealed} to the decision maker.

Let $\bG_t\in\mathbb{R}^{K\times K}$  be the adjacent matrix that represents the deterministic graph feedback structure $G_t$. Let $\bG_t(i,j)$ be the element at the $i$-th row and $j$-th column of the matrix. Then $\bG_t(i,j)=1$ if there exists an edge $(i,j)\in\cE_t$ and $\bG_t(i,j)=0$ otherwise. Note that we assume $\bG_t(i,i)=1$ for any $i\in\cK$. 
\begin{definition}\emph{(Clique cover number)
A \emph{clique} of a graph $G=(\cK,\cE)$ is a subset $S\subseteq \cK$ such that the sub-graph formed by $S$ and $\cE$ is a complete graph.
A \emph{clique cover} of a graph $G=(\cK,\cE)$ is a partition of $\cK$, denoted by $\mathcal{C}$, such that $S$ is a clique for each $S\in\mathcal{C}$. 
The cardinality of the smallest clique cover is called the \emph{clique cover number}, which is denoted by $\chi(G)$.
}
\end{definition}

\begin{definition}\emph{(Independence number)
An \emph{independent set} of a graph $G=(\cK,\cE)$ is a subset $S\subseteq \cK$ such that no two $i,j\in\cK$ are connected by an edge in $\cE$. The cardinality of a largest independent set is the \emph{independence number} of $G$, denoted by $\beta_0(G)$. 
}
\end{definition}
Note that the independence number of a directed graph is equivalent to that of the undirected graph by ignoring arc orientation. We can also lift the notion of independence number of an undirected graph to directed graph through the notion of maximum acyclic subgraphs.

\begin{definition}\emph{(Maximum acyclic subgraphs)
An \emph{acyclic subgraph} of $G=(\cK,\cE)$ is any graph $G'=(\cK',\cE')$ such that $\cK'\subseteq \cK$, and $\cE'=\cE\cap\left(\cK'\times\cK'\right)$, with no directed cycles. The cardinality of the largest such $\cK'$ is the \emph{maximum acyclic subgraphs} number, denoted by $mas(G)$.
}
\end{definition}
Note that $mas(G)\geq \beta_0(G)$ in general. The equality holds when the graph $G$ is undirected.
In this work, we slightly abuse the notation of the above graph numbers and use $\chi(G_t)$ and $\chi(\bG_t)$ interchangeably since $\bG_t$ fully characterizes the graph structure $G_t$.

%
%

\subsection{Randomized Policies}
We define all random variables with respect to a probability space $(\Omega,\mathcal{F},\mathbb{P})$. Consider the filtration $(\mathcal{F}_t)_{t\in\mathbb{N}}$ such that $\mathcal{F}_t\subseteq \mathcal{F}$ is the $\sigma$-algebra generated by the observation history $O_{t-1}$. The observation history $O_t$ includes all decisions, rewards and side observations from time $1$ to time $t$. For each time $t$, the decision maker chooses an action based on the history $O_{t-1}$ and possibly some randomness. Any policy of the decision maker can be viewed as a \emph{randomized policy} $\bpi$, which is an $\mathcal{F}_t$-adapted sequence $(\bpi_t)_{t\in\mathbb{N}}$. For each time $t$, the decision maker chooses an action randomly according to $\bpi_t(\cdot)=\mathbb{P}(A_t=\cdot|\mathcal{F}_t)$, which is a probability distribution over $\cK$. Let $\mathbb{E}[R(T,\bpi)]$ be the Bayesian regret defined by~(\ref{eqn:regret}) when the decisions $(A_1,\ldots,A_T)$ are chosen according to $\bpi$.

Uncertainty about $p^*$ induces uncertainty about the true optimal action $A^*$, which is described by a prior distribution $\balpha_1$ of $A^*$. Let $\balpha_t$ be the posterior distribution of $A^*$ given the history $O_{t-1}$, i.e., $\balpha_t(\cdot)=\mathbb{P}(A^*=\cdot|\mathcal{F}_t)$. Then, $\balpha_{t+1}$ can be updated by Bayes rule given $\balpha_{t}$, decision $A_t$, reward $Y_{t,A_t}$ and side observations. The \emph{Shannon entropy} of $\balpha_t$ is defined as $H(\balpha_t)\triangleq-\sum_{i\in\cK}\balpha_t(i)\log(\balpha_t(i))$. We slightly abuse the notion of $\bpi_t$ and $\balpha_t$ such that they represent distributions (or functions) over the finite set $\cK$ as well as vectors in a simplex $\mathcal{S}\subset\mathbb{R}^K$. Note that $\mathcal{S}=\{\bpi\in\mathbb{R}^K|\sum_{i=1}^K\bpi(i)=1, \bpi(i)\geq 0, \forall i\in\cK\}$.

Let $\bdelta_t$ be the instantaneous regret vector such that the $i$-th coordinate, $\bdelta_t(i)\triangleq\mathbb{E}[Y_{t,A^*}-Y_{t,i}|\mathcal{F}_t]$, is the expected regret of playing action $i$ at time $t$. Let $\bg_t$ be the information gain vector such that the $i$-th coordinate, $\bg_t(i)=\mathbb{E}[H(\balpha_t)-H(\balpha_{t+1})|\mathcal{F}_t,A_t=i]$, is the expected information gain of playing action $i$ at time $t$. Note that the information gain of playing action $i$ consists of that of observing the reward $Y_{t,i}$ and possibly some side observations. We define the information gain of observing action $a$ (i.e., $Y_{t,a}$) as $\bh_t(a)\triangleq I_t(A^*;Y_{t,a})$, which is the \emph{mutual information} under the posterior distribution between random variables $A^*$ and $Y_{t,a}$. Let $D(\cdot ||\cdot)$ be the \emph{Kullback-Leibler} divergence between two distributions\footnote{If $P$ is absolutely continuous with respect to $Q$, then $D(P||Q)=\int \log\left(\frac{\text{d}P}{\text{d}Q}\right)\text{d}P$, where $\frac{\text{d}P}{\text{d}Q}$ is the Radon-Nikodym derivative of $P$ w.r.t. $Q$.}. By the definition of mutual information, we have that $I_t(A^*;Y_{t,a})\triangleq$
\begin{equation}
D(\mathbb{P}((A^*,Y_{t,a})\in\cdot|\mathcal{F}_t)||\mathbb{P}(A^*\in\cdot|\mathcal{F}_t)\mathbb{P}(Y_{t,a}\in\cdot|\mathcal{F}_t)).
\end{equation}

At each time $t$, a randomized policy updates $\balpha_t$, and makes a decision according to a sampling distribution $\bpi_t$. For any randomized policy, we define the \emph{information ratio}~(\cite{russo2016information}) of sampling distribution $\bpi_t$ at time $t$ as 
\begin{equation}
\Psi_t(\bpi_t)\triangleq{(\bpi_t^T\bdelta_t)^2}/{(\bpi_t^T\bg_t)}.
\end{equation}
Note that $\bpi_t^T\bdelta_t$ is the expected instantaneous regret of the sampling distribution $\bpi_t$, and $\bpi_t^T\bg_t$ is the expected information gain of the sampling distribution $\bpi_t$. So the information ratio $\Psi_t(\bpi_t)$ measures the ``energy'' cost (which is the square of the expected instantaneous regret) per bit of information acquired. 

\subsection{Thompson Sampling}
Thompson Sampling algorithm simply samples actions according to the posterior probability that they are optimal. In particular, actions are chosen randomly at time $t$ according to the sampling distribution $\bpi_t=\balpha_t$. This conceptually elegant policy can be efficiently implemented. Consider the case where $\mathcal{P}=\{p_\theta\}_{\theta\in\Theta}$ is some parametric family of distributions. The true reward distribution $p^*$ is indexed by $\theta^*\in\Theta$ in the sense that $p^*=p_{\theta^*}$. Practical implementations of Thompson Sampling consist of two steps. First, an index $\hat{\theta}_t\sim\mathbb{P}(\theta^*\in\cdot|\cF_t)$ is sampled from the posterior distribution. Then, the algorithm selects the action $A_t=\arg\max_{a\in\cK}\mathbb{E}\left[Y_{t,a}|\theta^*=\hat{\theta}_t\right]$ that would be optimal if the sampled parameter were the true parameter. Given the observation of playing $A_t$, the posterior distribution is updated by Bayes' rule.

\section{Vanilla Thompson Sampling}
\begin{algorithm}[tb]
\caption{TS-N algorithm}
\label{alg:TSN}
\begin{algorithmic}
\REQUIRE time horizon $T$
\FOR{$t$ {\bfseries from} $1$ {\bfseries to} $T$}
\STATE{{\bf Updating statistics:} compute $\balpha_t$ accordingly.}
\STATE{{\bf Generating policy:} $\bpi_t=\balpha_t$.}
\STATE{{\bf Sampling:} sample $A_t$ according to $\bpi_t$, play action $A_t$ and receive reward $Y_{t,A_t}$.}
\STATE{{\bf Observations:} observe $Y_{t,a}$ if $(A_t,a)\in\cE_t$, where $G_t=(\cK,\cE_t)$ is the latent graph.}
\ENDFOR
\end{algorithmic}
\end{algorithm}
In this section, we show that a vanilla Thompson Sampling algorithm, TS-N as shown in Algorithm \ref{alg:TSN}, obtains optimal regret (within a logarithmic factor) in the undirected setting. A slightly weaker regret bound in the directed setting is also presented.

TS-N is the Thompson Sampling algorithm for graphical bandits such that $\bpi_t=\balpha_t$, where $\balpha_t$ is updated based on all the observations available, without additional modifications. It naturally keeps the information ratio $\Psi_t(\bpi_t)$ bounded as well as balances between having low expected instantaneous regret (a.k.a. exploitation) and obtaining knowledge about the optimal action (a.k.a. exploration). If the information ratio is bounded, then the expected regret is bounded in terms of the maximum amount of information one could expect to acquire, which is at most the entropy of the prior distribution of $A^*$, i.e., $H(\balpha_1)$. First, we bound the information ratio of TS-N in terms of the key quantity
\begin{equation}
Q_t(\bpi_t)=\sum_{i\in\cK}\frac{\bpi_t(i)}{\sum_{j:j\overset{t}{\to} i}\bpi_t(j)}.
\end{equation}
Note that $j\overset{t}{\to} i$ represents an arc $(j,i)$ in graph $G_t$.
\begin{proposition}\label{prop:TSNinforatio}
If $\bpi_t=\balpha_t$, then the information ratio satisfies $\Psi_t(\bpi_t)\leq \frac{1}{2}Q_t(\bpi_t)$ almost surely.
\end{proposition}

Proposition~\ref{prop:TSNinforatio} is a tight bound for the information ratio of the vanilla Thompson Sampling~(\cite{thompson}). If the graph is empty (i.e., there is no edges in the graph), then the quantity $Q_t(\bpi_t)$ equals to $K$. If the graph is complete, then the quantity $Q_t(\bpi_t)$ equals to $1$. These recover the information ratio bounds shown by \cite{russo2016information}. Also, the quantity $Q_t(\bpi_t)$ is upper bounded by clique cover number $\chi(G_t)$ as one can separate the sum by cliques and dropping the weights out of the clique. This recovers the information ratio bound shown by \cite{liu2017information}. Proposition~\ref{prop:TSNinforatio} allows us to show a tighter regret bound of Thompson Sampling for graphical bandits.
Next, we bound the regret of TS-N in terms of the quantity $Q_t(\bpi_t)$.
\begin{theorem}\label{thm:TSN}
The regret of TS-N satisfies
\begin{equation}
\mathbb{E}[R(T,\bpi)]\leq\sqrt{\frac{1}{2}\sum_{t=1}^T\mathbb{E}[Q_t(\balpha_t)]H(\balpha_1)}.
\end{equation}
\end{theorem}
Note that the entropy is bounded, i.e., $H(\balpha_1)\leq\log K$.
It has been shown by \cite{mannor,alon2014nonstochastic} that the quantity $Q_t(\bpi_t)$ is related to the graph numbers irrespective of the choice of the distribution $\bpi_t$. The following graph-theoretic result shows that the quantity $Q_t(\bpi_t)$ is bounded by the independence number of the latent graph if the graph is undirected.
\begin{lemma}\label{lem:undirected}
\emph{(Lemma 3 in~\cite{mannor})} Let $G=(\cK,\cE)$ be an undirected graph. For any distribution $\bpi$ over $\cK$,
\begin{equation}
\sum_{i\in\cK}\frac{\bpi(i)}{\sum_{j:j\overset{t}{\to} i}\bpi(j)}\leq \beta_0(G).
\end{equation}
\end{lemma}
The following regret result of TS-N follows immediately from Theorem~\ref{thm:TSN} and Lemma~\ref{lem:undirected}.
\begin{corollary}\label{cor:TSNundirected}
In the undirected setting, the regret of TS-N satisfies
\begin{equation}
\mathbb{E}[R(T,\bpi)]\leq\sqrt{\frac{1}{2}\sum_{t=1}^T\beta_0(G_t)H(\balpha_1)}.
\end{equation}
\end{corollary}
As far as we know, this is the best regret bound for graphical bandits without the graphs. First, an information-theoretic lower bound of graphical bandits has been shown by~\cite{mannor,alon2014nonstochastic} to be $\Omega\left(\sqrt{\beta_0(G)T}\right)$. So Corollary~\ref{cor:TSNundirected} shows that TS-N obtains the optimal regret (within a logarithmic factor) in the undirected setting. Moreover, the bound proven in Corollary~\ref{cor:TSNundirected} is tighter than the $O\left(\sqrt{\chi(G)TH(\balpha_1)}\right)$ bound of TS-N shown by~\cite{liu2017information} as $\beta_0(G)\leq\chi(G)$. At last, Corollary~\ref{cor:TSNundirected} shows that TS-N enjoys better regret bound than Cohen's Algorithm 1 developed by~\cite{Cohen2016} in the undirected setting, both of which do not require the knowledge of the feedback graphs.

We now turn to the directed setting. The following graph-theoretic result shows that the quantity $Q_t(\bpi_t)$ is upper-bounded by the maximum acyclic subgraph number if the graph is directed.
\begin{lemma}\label{lem:directed}
\emph{(Lemma 10 in~\cite{alon2014nonstochastic})} Let $G=(\cK,\cE)$ be a directed graph. For any distribution $\bpi$ over $\cK$,
\begin{equation}
\sum_{i\in\cK}\frac{\bpi(i)}{\sum_{j:j\overset{t}{\to} i}\bpi(j)}\leq mas(G).
\end{equation}
\end{lemma}
The following regret result of TS-N follows immediately from Theorem~\ref{thm:TSN} and Lemma~\ref{lem:directed}.
\begin{corollary}\label{cor:TSNdirected}
In the directed setting, the regret of TS-N satisfies
\begin{equation}
\mathbb{E}[R(T,\bpi)]\leq\sqrt{\frac{1}{2}\sum_{t=1}^Tmas(G_t)H(\balpha_1)}.
\end{equation}
\end{corollary}
The bound proven in Corollary~\ref{cor:TSNdirected} is tighter than the $O\left(\sqrt{\chi(G)TH(\balpha_1)}\right)$ bound of TS-N shown by~\cite{liu2017information} since $mas(G)\leq\chi(G)$. However, there is a gap between the lower bound and the upper bound shown in Corollary~\ref{cor:TSNdirected}. Though $\beta_0(G)=mas(G)$ when the graph is undirected, the gap between them can be large in general directed graphs. For example, consider a directed graph $G_0=(\cK,\cE)$ such that arc $(i,j)\in\cE$ if and only if $i\leq j$. It is clear that $\beta_0(G_0)=1$ and $mas(G_0)=K$. This leads us to consider a more sophisticated randomized policy. In the next section, we show that a modified Thompson Sampling algorithm results in an optimal (within a logarithmic factor) regret bound in the general setting.

\begin{remark}
The bounds proven in Corollary~\ref{cor:TSNundirected} and~\ref{cor:TSNdirected} hold for IDS-N and IDSN-LP developed by~\cite{liu2017information} since the information ratio of IDS-N and IDSN-LP are bounded by the information ratio of TS-N almost surely. So our results also provide tighter bounds for IDS-N and IDSN-LP algorithms. Note that IDS-N and IDSN-LP algorithms require prior knowledge of the feedback graphs. 
\end{remark}

\section{Thompson Sampling with Exploration}

\begin{algorithm}[tb]
\caption{TS-U algorithm}
\label{alg:TSU}
\begin{algorithmic}
\REQUIRE time horizon $T$ and parameter $\epsilon\in[0,1]$
\FOR{$t$ {\bfseries from} $1$ {\bfseries to} $T$}
\STATE{{\bf Updating statistics:} compute $\balpha_t$ accordingly.}
\STATE{{\bf Generating policy:} $\bpi_t=(1-\epsilon)\balpha_t+\epsilon/K$.}
\STATE{{\bf Sampling:} sample $A_t$ according to $\bpi_t$, play action $A_t$ and receive reward $Y_{t,A_t}$.}
\STATE{{\bf Observations:} observe $Y_{t,a}$ if $(A_t,a)\in\cE_t$, where $G_t=(\cK,\cE_t)$ is the latent graph.}
\ENDFOR
\end{algorithmic}
\end{algorithm}
In this section, we show that a mixture of Thompson Sampling and uniform sampling, TS-U as shown in Algorithm~\ref{alg:TSU}, obtains the optimal regret within a logarithmic factor in the directed setting. 

TS-U algorithm is a variant of Thompson Sampling with explicit exploration that allows the algorithm to explore some suboptimal actions with large out-degrees. The collected side observations from the uniform sampling allows us to capture the latent graph information, thus yielding a regret bound in terms of the independence number.

As shown in Algorithm~\ref{alg:TSU}, TS-U is a randomized policy such that $\bpi_t=(1-\epsilon)\balpha_t+\epsilon/K$ for some parameter $\epsilon\in[0,1]$. The implementation and computation of TS-U is quite efficient. At each time $t$, TS-U algorithm plays vanilla Thompson Sampling with probability $1-\epsilon$ and plays uniform sampling with probability $\epsilon$. While the expected information gain diminishes, the expected instantaneous regret is bounded away from zero due to uniform sampling. Thus, the classical analysis of bounding the information ratio $\Psi_t(\bpi_t)$ does not work any more. Fortunately, the linear form of $\bpi_t$ allows us to bound the regret of TS-U by the regret due to uniform sampling plus the regret from Thompson Sampling, as shown in Theorem~\ref{thm:TSU}. Indeed, our techniques work for any variant of Thompson Sampling that is a linear combination of $\balpha_t$ and some other distributions.

\begin{theorem}\label{thm:TSU}
The regret of TS-U satisfies
\begin{equation}
\mathbb{E}[R(T,\bpi)]\leq\epsilon T+\sqrt{\frac{1}{2}\sum_{t=1}^T\mathbb{E}[Q_t(\bpi_t)]H(\balpha_1)}.
\end{equation}
\end{theorem}

Theorem~\ref{thm:TSU} shows that the regret of TS-U consists of two parts, the regret from the uniform sampling and the regret from the Thompson Sampling. Note that the regret from the Thompson Sampling takes into account the information gain from the uniform sampling as the term $Q_t(\bpi_t)$ depends on $\bpi_t$ rather than $\balpha_t$. This allows us to use the following graph-theoretic result to bound the term $Q_t(\bpi_t)$, thus the regret of TS-U, by the independence number of the latent graph.

\begin{lemma}\label{lem:TSU}
\emph{(Lemma 5 in \cite{alon2015online})} Let $G=(\cK,\cE)$ be a directed graph. For any distribution $\bpi$ over $\cK$ such that $\bpi(i)\geq\eta$ for all $i\in\cK$ for some constant $0<\eta<0.5$. Then
\begin{equation}
\sum_{i\in\cK}\frac{\bpi(i)}{\sum_{j:j\overset{t}{\to} i}\bpi(j)}\leq 4\beta_0(G)\log\left(\frac{4K}{\beta_0(G)\eta}\right).
\end{equation}
\end{lemma}
Lemma~\ref{lem:TSU} shows that the quantity $Q_t(\bpi_t)$ can be bounded by the independence number of the directed graph if $\bpi_t$ is bounded away from zero. The uniform sampling part of TS-U allows the sampling distribution to satisfy this condition. By Theorem~\ref{thm:TSU} and taking $\eta=\epsilon/K$ in Lemma~\ref{lem:TSU}, we have the following result.
\begin{corollary}\label{cor:TSU}
If $\epsilon=1/\sqrt{T}$, then the regret of TS-U satisfies
\begin{equation}
\mathbb{E}[R(T,\bpi)]=O\left(\sqrt{\log(KT)\sum_{t=1}^T\beta_0(G_t)H(\balpha_1)}\right).
\end{equation}
\end{corollary}

Comparing the regret bound in Corollary~\ref{cor:TSU} to the lower bound, $\Omega(\sqrt{\beta_0(G)T})$, the TS-U algorithm obtains the optimal regret within a logarithmic factor in the general setting. Moreover, Corollary~\ref{cor:TSU} shows that TS-U enjoys a sharper (by a logarithmic factor) regret bound than Cohen's Algorithm 1 developed by~\cite{Cohen2016} in the directed setting, both of which do not require the knowledge of the feedback graphs. As far as we know, this is the best-known regret bound for graphical bandits without the graphs. Finally, note that a comparison between Corollary~\ref{cor:TSNundirected} and Corollary~\ref{cor:TSU} reveals that a symmetric observation system (i.e., undirected feedback graphs) enjoys better regret than an asymmetric observation system as the regret bound of TS-N is sharper by a logarithmic factor than the bound in Corollary~\ref{cor:TSU} in the undirected setting.

\begin{remark}\label{remark:epsilon}
We present TS-U algorithm with fixed exploration rate $\epsilon$ for simplicity. It is easy to verify that the regret result of TS-U still holds if one uses some appropriate decreasing exploration rate sequence $\{\epsilon_t\}$. For example, when $\epsilon_t=1/t$, Theorem~\ref{thm:TSU} holds by replacing the term $\epsilon T$ with $\log T$. Then the regret result of the corresponding TS-U algorithm follows. In practice, we recommend practitioners to use decreasing exploration rates.
\end{remark}

\begin{remark}
In the informed setting (i.e., when the feedback graph are revealed to the decision maker before the decisions), one may propose a variant of TS-U such that it restricts the exploration set to the dominating set of the feedback graph. In other words, one may replace uniform sampling over all the actions with uniform sampling over only the dominating set. The regret bound in Corollary~\ref{cor:TSU} still holds by a variant of Lemma~\ref{lem:TSU} shown in~\cite{alon2014nonstochastic}.
\end{remark}

\section{Numerical Results}

\begin{figure}[t]
\centering
\begin{subfigure}[b]{0.45\textwidth}
    \includegraphics[width=\textwidth]{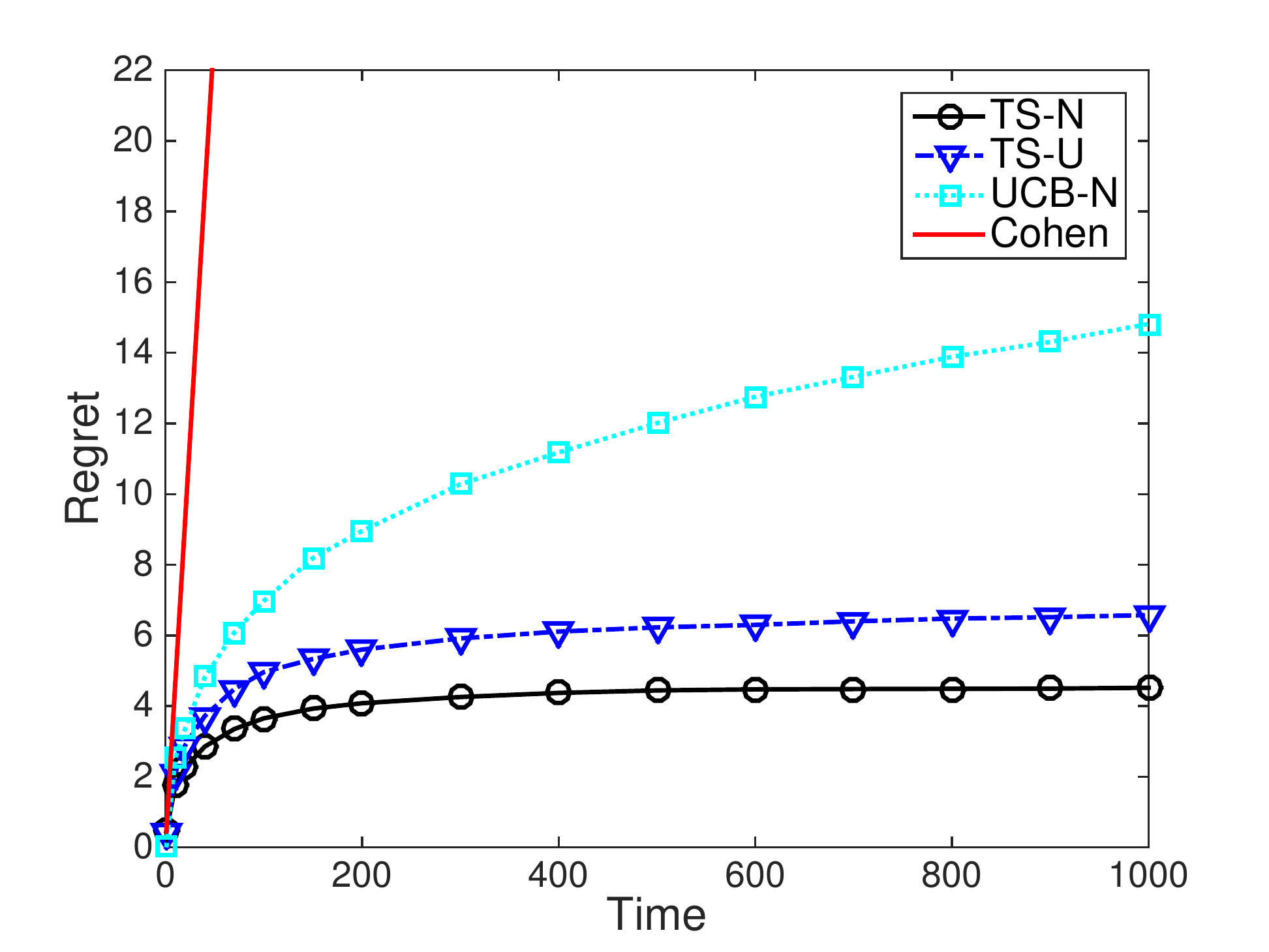}
     \caption{Time-invariant graphs}\label{subfig:undirinvar}
\end{subfigure}
\begin{subfigure}[b]{0.45\textwidth}
    \includegraphics[width=\textwidth]{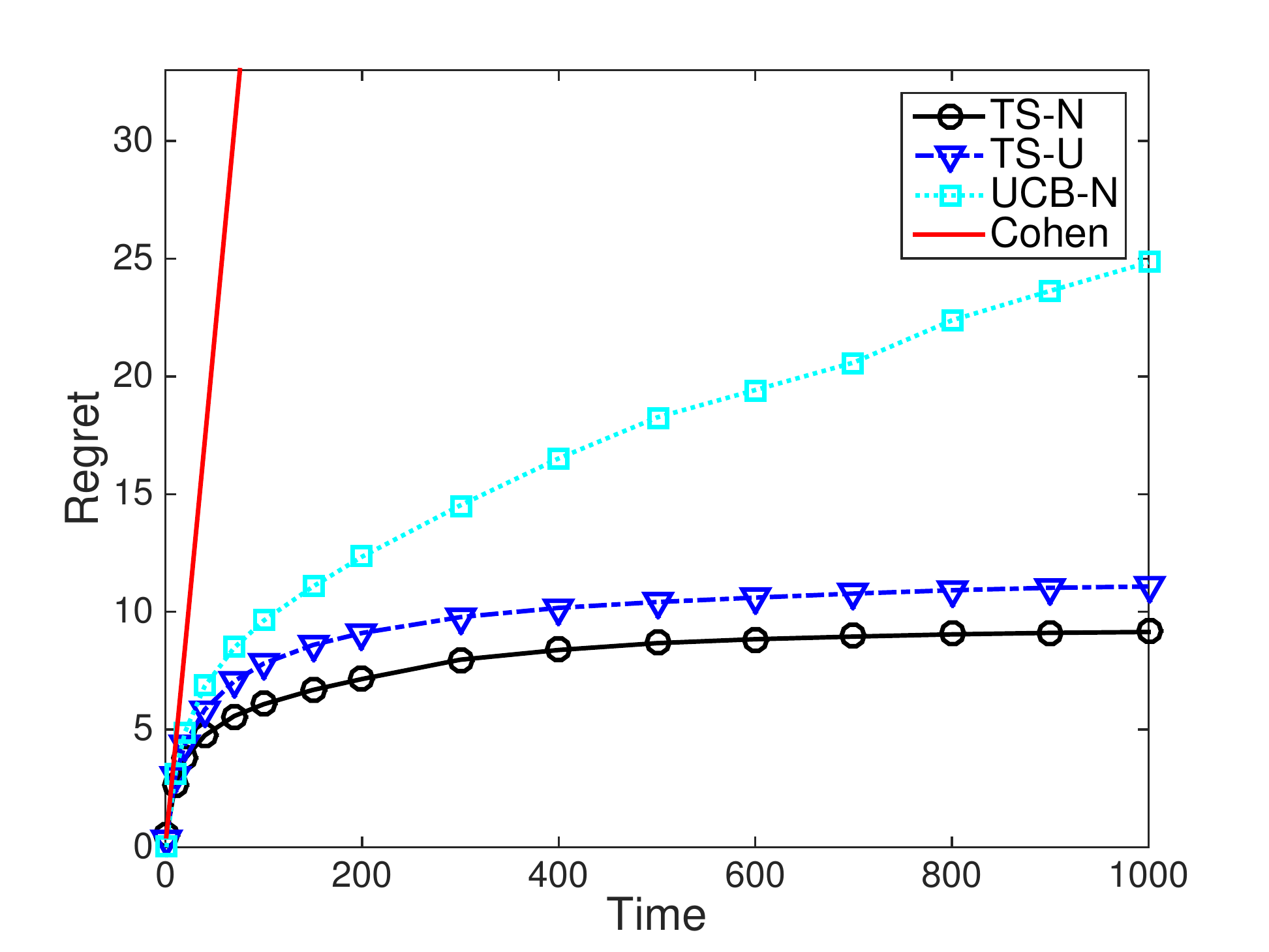}
     \caption{Time-variant graphs}\label{subfig:undirvar}
\end{subfigure}
\caption{Regret comparison under undirected graph feedback}
\label{fig:undir}
\end{figure}

\begin{figure}[t]
\centering
\begin{subfigure}[b]{0.45\textwidth}
    \includegraphics[width=\textwidth]{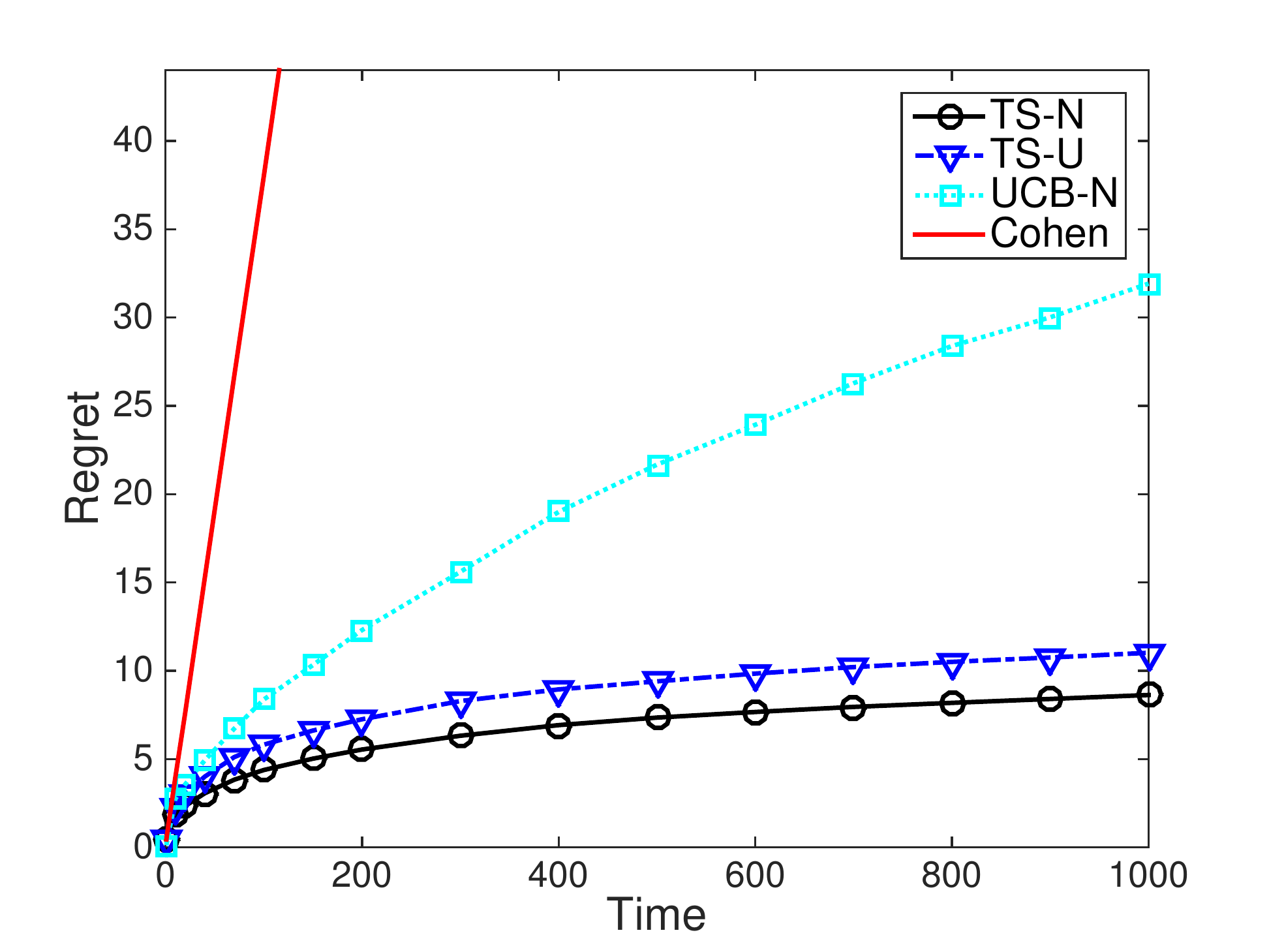}
     \caption{Time-invariant graphs}\label{subfig:dirinvar}
\end{subfigure}
\begin{subfigure}[b]{0.45\textwidth}
    \includegraphics[width=\textwidth]{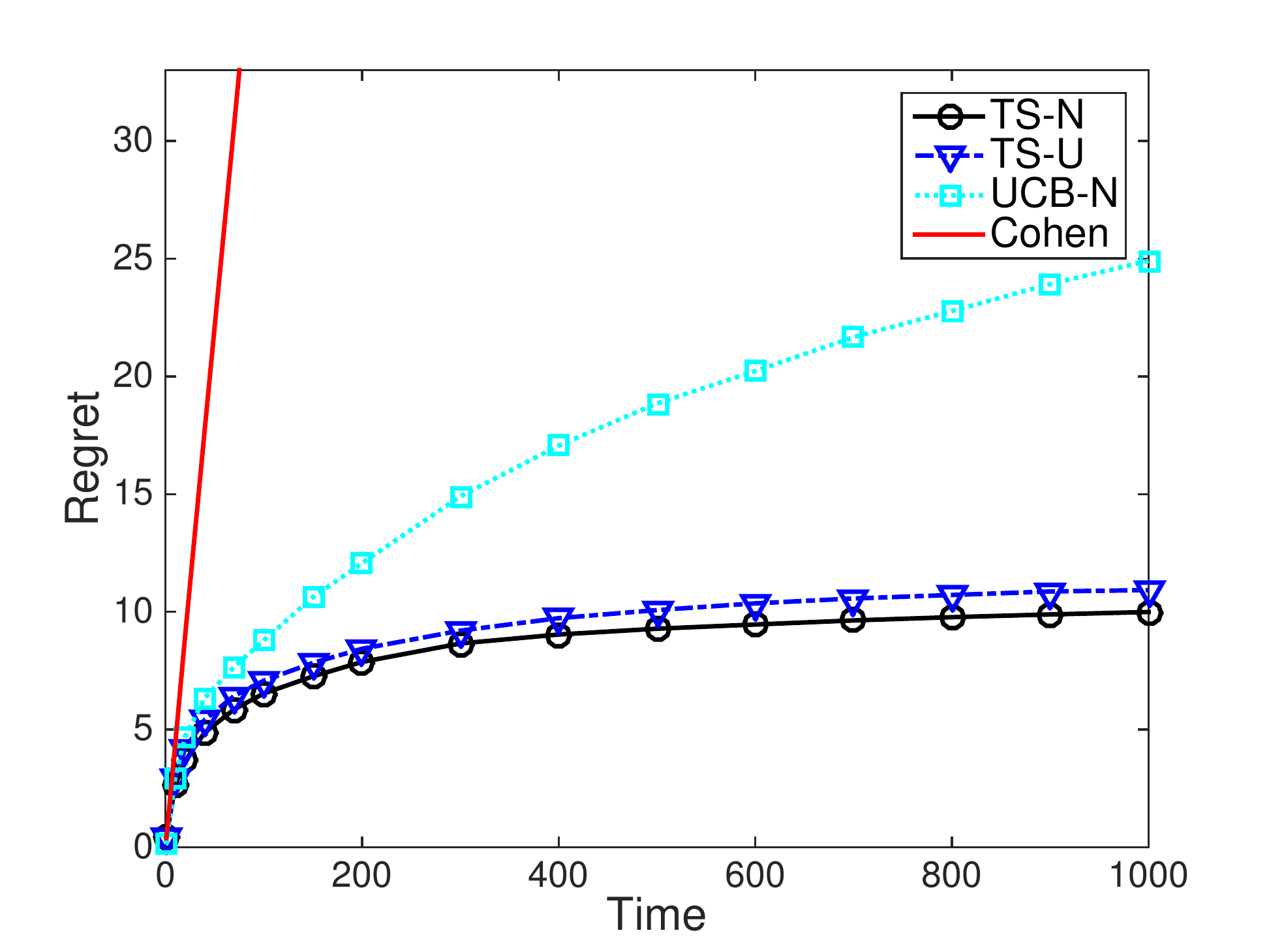}
     \caption{Time-variant graphs}\label{subfig:dirvar}
\end{subfigure}
\caption{Regret comparison under directed graph feedback}
\label{fig:dir}
\end{figure}

This section presents numerical results from experiments that evaluate the effectiveness of Thompson Sampling based policies in comparison to UCB-N and Cohen's algorithm. We consider the classical Beta-Bernoulli bandit problem with independent actions. The reward of each action $i$ is a Bernoulli$(\mu_i)$ random variable and $\mu_i$ is independently drawn from Beta$(1,1)$. The implementations of TS-N and TS-U are shown in Algorithms~\ref{alg:TSNBernoulli} and~\ref{alg:TSUBernoulli}. In the experiment, we set $K=5$, $T=1000$ and $\epsilon_t=1/t$ as suggested by Remark~\ref{remark:epsilon}. All the regret results are averaged over $1000$ trials. 

\begin{figure}[t]
  \centering
    \includegraphics[width=0.2\textwidth]{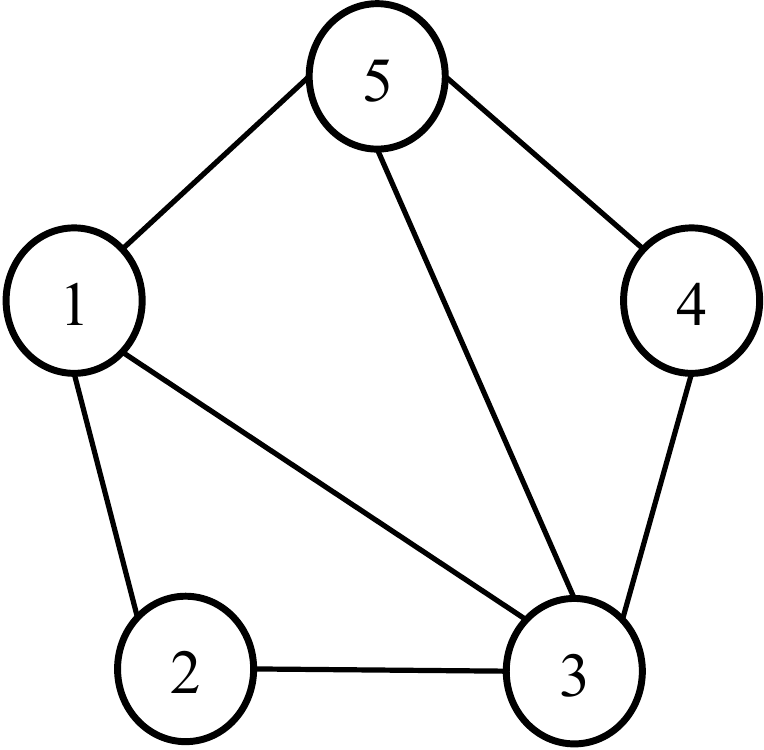}
     \caption{Graph structure for the experiment under time-invariant and undirected graph}
     \label{fig:undirinvargraph}
\end{figure}

\begin{figure}[t]
  \centering
    \includegraphics[width=0.2\textwidth]{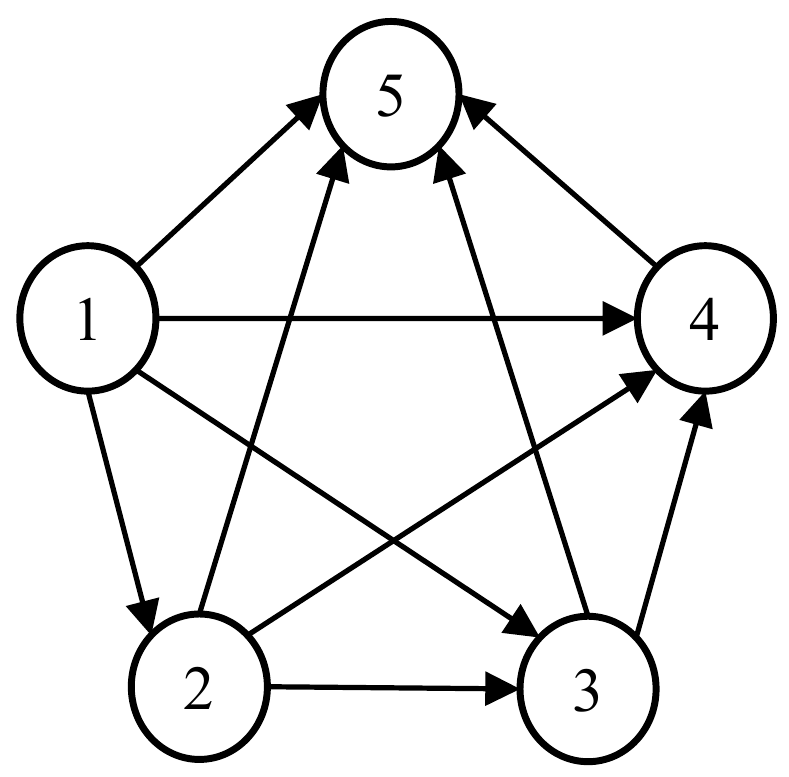}
     \caption{Graph structure for the experiment under time-invariant and directed graph}
     \label{fig:dirinvargraph}
\end{figure}

Figure~\ref{fig:undir} presents the cumulative regret results under undirected graph feedback. For the time-invariant case shown in Figure~\ref{subfig:undirinvar}, we use a graph with 2 cliques, presented in Figure~\ref{fig:undirinvargraph}. Figure~\ref{fig:dir} presents the cumulative regret results under directed graph feedback. For the time-invariant case shown in Figure~\ref{subfig:dirinvar}, we use the graph $G=(\cK,\cE)$ such that $(i,j)\in\cE$ if and only $i\leq j$, presented in Figure~\ref{fig:dirinvargraph}. For the time-variant cases shown in Figures~\ref{subfig:undirvar} and~\ref{subfig:dirvar}, the sequences of graphs are generated by the Erd{\H{o}}s-R{\'e}nyi model with parameter $p_t$\footnote{For each time $t$ and each pair $(i,j)$, $(i,j)\in\cE_t$ with probability $p_t$} drawn from the uniform distribution over $[0,0.2]$.

We find that TS-N and TS-U outperform the alternative algorithms, which is consistent with the empirical observation in the bandit feedback setting~(\cite{chapelle2011empirical}). However, TS-N has better empirical performance in the tested settings even though we have proven a better regret bound for TS-U under directed feedback graphs. The average regrets of Cohen's algorithm are dramatically larger than that of Thompson Sampling based policies. For this reason, parts of Cohen's algorithm are omitted from Figures~\ref{fig:undir} and~\ref{fig:dir}.

\begin{algorithm}[tb]
\caption{TS-N (Bernoulli case)}
\label{alg:TSNBernoulli}
\begin{algorithmic}
\REQUIRE time horizon $T$
\STATE{For each arm $i$, set $S_i=1$ and $F_i=1$}
\FOR{$t$ {\bfseries from} $1$ {\bfseries to} $T$}
\STATE{For each arm $i$, sample $\theta_i$ from Beta$(S_i,F_i)$.}
\STATE{Play action $A_t=\arg\max_{i\in\cK}\theta_i$.}
\FOR{all $a\in\cK$ such that $(A_t,a)\in\cE_t$}
\STATE{$S_a=S_a+Y_{t,a}$ and $F_a=F_a+1-Y_{t,a}$.}
\ENDFOR
\ENDFOR
\end{algorithmic}
\end{algorithm}

\begin{algorithm}[tb]
\caption{TS-U (Bernoulli case)}
\label{alg:TSUBernoulli}
\begin{algorithmic}
\REQUIRE time horizon $T$ and $\{\epsilon_t\}_{t\geq 1}$
\STATE{For each arm $i$, set $S_i=1$ and $F_i=1$}
\FOR{$t$ {\bfseries from} $1$ {\bfseries to} $T$}
\STATE{Sample $\beta_t$ from uniform distribution over $[0,1]$.}
\IF{$\beta_t<\epsilon_t$}
\STATE{Play action $A_t$ drawn uniformly from $\cK$.}
\ELSE
\STATE{For each arm $i$, sample $\theta_i$ from Beta$(S_i,F_i)$.}
\STATE{Play action $A_t=\arg\max_{i\in\cK}\theta_i$.}
\ENDIF
\FOR{all $a\in\cK$ such that $(A_t,a)\in\cE_t$}
\STATE{$S_a=S_a+Y_{t,a}$ and $F_a=F_a+1-Y_{t,a}$.}
\ENDFOR
\ENDFOR
\end{algorithmic}
\end{algorithm}

\section{Conclusion}
We have provided regret analysis of Thompson Sampling for graphical bandits without knowing the feedback graphs at any time. We show that the regret of TS-N is bounded by $O\left(\sqrt{mas(G)T\log K}\right)$ in the general setting. In the undirected setting, $mas(G)=\beta_0(G)$, and the resulting regret bound is optimal up to a logarithmic factor. As far as we know, this is the first result that shows that Thompson Sampling, even without the knowledge of the graph, can attain the optimal regret in the graphical bandits.  As a byproduct, our analysis for TS-N provide improved regret bounds for information directed sampling algorithms (IDS-N and IDSN-LP algorithms) proposed by~\cite{liu2017information} in the informed setting. 

We have proposed a variant of Thompson Sampling, TS-U, that mixes Thompson Sampling with uniform sampling. This modification allows the algorithm to capture the graph structure and obtain $O\left(\sqrt{\beta_0(G)T\log K\log (KT)}\right)$ regret bound in the directed setting, which is optimal within a logarithmic factor. Our results offer a recipe for practitioners to choose algorithms for graphical bandits without knowledge of the graphs. If the latent graphs are known to be undirected, one can choose TS-N for the best regret guarantee. Otherwise, TS-U is the choice with the best guarantee in the general (directed) setting.

\newpage
\bibliographystyle{named}
\bibliography{refs,ref}

\newpage
\appendix

\section{Proof of Proposition~\ref{prop:TSNinforatio}}
As shorthand, we let $d(a,a^*)=D(\mathbb{P}(Y_{t,a}\in\cdot|\cF_t,A^*=a^*)||\mathbb{P}(Y_{t,a}\in\cdot|\cF_t))$ and $x(a)=\sqrt{d(a,a)}$.
By the definition of the instantaneous regret, we have that
\begin{align}
\bdelta_t^T\balpha_t&=\sum_{a\in\cK}\balpha_t(a)\mathbb{E}[Y_{t,A^*}-Y_{t,a}|\mathcal{F}_t]\\
&\overset{(a)}{=}\sum_{a\in\cK}\balpha_t(a)\left(\mathbb{E}[Y_{t,A^*}|\cF_t]-\mathbb{E}[Y_{t,a}|\mathcal{F}_t]\right)\\
&\overset{}{=}\mathbb{E}[Y_{t,A^*}|\cF_t]-\sum_{a\in\cK}\balpha_t(a)\mathbb{E}[Y_{t,a}|\mathcal{F}_t]\\\nonumber
&\overset{(b)}{=}\sum_{a\in\cK}\balpha_t(a)\left(\mathbb{E}[Y_{t,a}|\cF_t,A^*=a]-\mathbb{E}[Y_{t,a}|\mathcal{F}_t]\right)\\
&\overset{(c)}{\leq}\sqrt{\frac{1}{2}}\sum_{a\in\cK}\balpha_t(a)\sqrt{d(a,a)}\\
&=\sqrt{\frac{1}{2}}\sum_{a\in\cK}\balpha_t(a)x(a)
,\label{eqn:instantregret}
\end{align}
where $(a)$ follows from the linearity of expectation, $(b)$ uses the law of total probability, $(c)$ follows from the Pinsker's inequality.

By the definition of the information gain of observing an action, we have that
\begin{align}\nonumber
\bg_t^T\balpha_t&\overset{(d)}{\geq}(\bG_t\bh_t)^T\balpha_t\\
&=\sum_{a\in\cK}\left(\sum_{a':a'\overset{t}{\to}a}\balpha_t(a')\right)I_t(A^*;Y_{t,a})\\\nonumber
&\overset{(e)}{=}\sum_{a\in\cK}\left(\sum_{a':a'\overset{t}{\to}a}\balpha_t(a')\right)\left(\sum_{a^*\in\cK}\balpha_t(a^*)d(a,a^*)\right)\\
&\overset{(f)}{\geq}\sum_{a\in\cK}\left(\sum_{a':a'\overset{t}{\to}a}\balpha_t(a')\right)\balpha_t(a)d(a,a)\\
&=\sum_{a\in\cK}\left(\sum_{a':a'\overset{t}{\to}a}\balpha_t(a')\right)\balpha_t(a)\left(x(a)\right)^2
,\label{eqn:instantgain}
\end{align}
where $(d)$ follows from Proposition 1 of \cite{liu2017information}, $(e)$ follows from the KL divergence form of mutual information and $(f)$ follows by dropping some nonnegative terms.

As shorthand, we let $z(a)=\frac{\sum_{a':a'\overset{t}{\to}a}\balpha_t(a')}{\balpha_t(a)}$. Now, we are ready to bound the information ration.
\begin{align}
(\bdelta_t^T&\balpha_t)^2\overset{(g)}{\leq}\frac{1}{2}\left(\sum_{a\in\cK}\balpha_t(a)x(a)\right)^2\\\nonumber
&=\frac{1}{2}\left(\sum_{a\in\cK}\frac{1}{\sqrt{z(a)}}\sqrt{z(a)}\balpha_t(a)x(a)\right)^2\\\nonumber
&\overset{(h)}{\leq}\frac{1}{2}\left(\sum_{a\in\cK}\frac{1}{z(a)}\right)\left(\sum_{a\in\cK}z(a)\left(\balpha_t(a)x(a)\right)^2\right)\\
&\overset{(i)}{\leq}\frac{1}{2}\left(\sum_{a\in\cK}\frac{1}{z(a)}\right)\bg_t^T\balpha_t
\end{align}
where $(g)$ follows from equation $(\ref{eqn:instantregret})$, $(h)$ follows from Cauchy-Schwartz inequality and $(i)$ follows from equation $(\ref{eqn:instantgain})$.

\section{Proof of Theorem~\ref{thm:TSN}}
First observe that the entropy bounds the expected cumulative information gain.
\begin{align}
&\mathbb{E}\sum_{t=1}^T \bg_t^T\bpi_t\\\nonumber
&=\mathbb{E}\sum_{t=1}^T\left(\sum_{i\in\cK}\bpi_t(i)\mathbb{E}[H(\balpha_t)-H(\balpha_{t+1})|\cF_t,A_t=i]\right)\\
&=\mathbb{E}\sum_{t=1}^T\mathbb{E}[H(\balpha_t)-H(\balpha_{t+1})|\cF_t]\\
&=\mathbb{E}\sum_{t=1}^T\left(H(\balpha_t)-H(\balpha_{t+1})\right)\\
&\leq H(\balpha_1),\label{eqn:entropybound}
\end{align}

Then, we bound the regret of TS-N.
\begin{align}\nonumber
\mathbb{E}[R(T,\bpi)]&=\mathbb{E}\sum_{t=1}^T\bdelta_t^T\bpi_t=\mathbb{E}\sum_{t=1}^T\frac{\bdelta_t^T\bpi_t}{\sqrt{\bg_t^T\bpi_t}}\sqrt{\bg_t^T\bpi_t}\\\nonumber
&\overset{(a)}{\leq}\sqrt{\mathbb{E}\sum_{t=1}^T\frac{(\bdelta_t^T\bpi_t)^2}{\bg_t^T\bpi_t}}\sqrt{\mathbb{E}\sum_{t=1}^T\bg_t^T\bpi_t}\\
&\overset{(b)}{\leq}\sqrt{\frac{1}{2}\sum_{t=1}^T\mathbb{E}[Q_t(\balpha_t)]H(\balpha_1)}
\end{align}
where $(a)$ follows from Holder's inequality and $(b)$ follows from Proposition \ref{prop:TSNinforatio} and equation $(\ref{eqn:entropybound})$.

\section{Proof of Theorem~\ref{thm:TSU}}
As shorthand, we let $d(a,a^*)=D(\mathbb{P}(Y_{t,a}\in\cdot|\cF_t,A^*=a^*)||\mathbb{P}(Y_{t,a}\in\cdot|\cF_t))$ and $x(a)=\sqrt{d(a,a)}$.
By the definition of the instantaneous regret, we have that
\begin{align}
&\bdelta_t^T\bpi_t=\sum_{a\in\cK}\bpi_t(a)\mathbb{E}[Y_{t,A^*}-Y_{t,a}|\mathcal{F}_t]\\
&\overset{(a)}{=}\sum_{a\in\cK}\bpi_t(a)\left(\mathbb{E}[Y_{t,A^*}|\cF_t]-\mathbb{E}[Y_{t,a}|\mathcal{F}_t]\right)\\
&\overset{}{=}\mathbb{E}[Y_{t,A^*}|\cF_t]-\sum_{a\in\cK}\bpi_t(a)\mathbb{E}[Y_{t,a}|\mathcal{F}_t]\\\nonumber
&\overset{(b)}{=}\sum_{a\in\cK}\balpha_t(a)\mathbb{E}[Y_{t,a}|\cF_t,A^*=a]-\sum_{a\in\cK}\bpi_t(a)\mathbb{E}[Y_{t,a}|\mathcal{F}_t]\\\nonumber
&\overset{(c)}{\leq}(1\!-\!\epsilon)\sum_{a\in\cK}\balpha_t(a)\left(\mathbb{E}[Y_{t,a}|\cF_t,A^*\!=\!a]\!-\!\mathbb{E}[Y_{t,a}|\mathcal{F}_t]\right)\!+\!\epsilon\\
&\overset{(d)}{\leq}(1-\epsilon)\sqrt{\frac{1}{2}}\sum_{a\in\cK}\left(\balpha_t(a)\sqrt{d(a,a)}\right)+\epsilon\\
&=(1-\epsilon)\sqrt{\frac{1}{2}}\sum_{a\in\cK}\left(\balpha_t(a)x(a)\right)+\epsilon
,\label{eqn:TSUinstantregret}
\end{align}
where $(a)$ follows from the linearity of expectation, $(b)$ uses the law of total probability, $(c)$ follows from the fact that $\bpi_t=(1-\epsilon)\balpha_t+\epsilon/K$ and the rewards are bounded by 1, $(d)$ follows from the Pinsker's inequality. Step $(c)$ allows us to decompose the regret into the regret from uniform sampling and the regret from Thompson Sampling. Thus, we can further relate the latter regret term to the expected information gain.

By the definition of the information gain of observing an action, we have that
\begin{align}\nonumber
\bg_t^T\bpi_t&\overset{(e)}{\geq}(\bG_t\bh_t)^T\bpi_t\\
&=\sum_{a\in\cK}\left(\sum_{a':a'\overset{t}{\to}a}\bpi_t(a')\right)I_t(A^*;Y_{t,a})\\\nonumber
&\overset{(f)}{=}\sum_{a\in\cK}\left(\sum_{a':a'\overset{t}{\to}a}\bpi_t(a')\right)\left(\sum_{a^*\in\cK}\balpha_t(a^*)d(a,a^*)\right)\\
&\overset{(g)}{\geq}\sum_{a\in\cK}\left(\sum_{a':a'\overset{t}{\to}a}\bpi_t(a')\right)\balpha_t(a)d(a,a)\\
&=\sum_{a\in\cK}\left(\sum_{a':a'\overset{t}{\to}a}\bpi_t(a')\right)\balpha_t(a)\left(x(a)\right)^2
,\label{eqn:TSUinstantgain}
\end{align}
where $(e)$ follows from Proposition 1 of \cite{liu2017information}, $(f)$ follows from the KL divergence form of mutual information and $(g)$ follows by dropping some nonnegative terms.

As shorthand, we let $\zeta(a)=\frac{\sum_{a':a'\overset{t}{\to}a}\bpi_t(a')}{\balpha_t(a)}$. Now we are ready to bound the first term in equation $(\ref{eqn:TSUinstantregret})$.
\begin{align}
\sum_{a\in\cK}&\balpha_t(a)x(a)=\sum_{a\in\cK}\frac{1}{\sqrt{\zeta(a)}}\sqrt{\zeta(a)}\balpha_t(a)x(a)\\
&\overset{(h)}{\leq}\sqrt{\sum_{a\in\cK}\frac{1}{\zeta(a)}}\sqrt{\sum_{a\in\cK}\zeta(a)(\balpha_t(a)x(a))^2}\\
&\overset{(i)}{\leq}\sqrt{\sum_{a\in\cK}\frac{\balpha_t(a)}{\sum_{a':a'\overset{t}{\to}a}\bpi_t(a')}}\sqrt{\bg_t^T\bpi_t}\\
&\overset{(j)}{\leq}\sqrt{\frac{1}{1-\epsilon}\sum_{a\in\cK}\frac{\bpi_t(a)}{\sum_{a':a'\overset{t}{\to}a}\bpi_t(a')}}\sqrt{\bg_t^T\bpi_t}\\
&=\sqrt{\frac{1}{1-\epsilon}Q_t(\bpi_t)}\sqrt{\bg_t^T\bpi_t}
,\label{eqn:TSUbridge}
\end{align}
where $(h)$ follows from Cauchy-Schwartz inequality, $(i)$ follows from equation $(\ref{eqn:TSUinstantgain})$ and $(j)$ follows from the fact that $\bpi_t\geq(1-\epsilon)\balpha_t$.

Now, we are ready to bound the regret.
\begin{align}
\mathbb{E}&[R(T,\bpi)]=\mathbb{E}\sum_{t=1}^T\bdelta_t^T\bpi_t\\\nonumber
&\overset{(k)}{\leq}\mathbb{E}\sum_{t=1}^T\left((1-\epsilon)\sqrt{\frac{1}{2}}\sum_{a\in\cK}\left(\balpha_t(a)x(a)\right)+\epsilon\right)\\\nonumber
&\overset{(l)}{\leq}\epsilon T+\sqrt{\frac{1}{2}}\mathbb{E}\sum_{t=1}^T\sqrt{(1-\epsilon)Q_t(\bpi_t)}\sqrt{\bg_t^T\bpi_t}\\\nonumber
&\overset{(m)}{\leq}\epsilon T+\sqrt{\frac{1}{2}\mathbb{E}\sum_{t=1}^TQ_t(\bpi_t)}\sqrt{\mathbb{E}\sum_{t=1}^T\bg_t^T\bpi_t}\\
&\overset{(n)}{\leq}\epsilon T+\sqrt{\frac{1}{2}\sum_{t=1}^T\mathbb{E}[Q_t(\bpi_t)]H(\balpha_1)},
\end{align}
where $(k)$ follows from equation $(\ref{eqn:TSUinstantregret})$, $(l)$ follows from equation $(\ref{eqn:TSUbridge})$, $(m)$ follows from Holder's inequality and $(n)$ follows from equation $(\ref{eqn:entropybound})$.

\end{document}